\documentclass{article}

\usepackage[final,nonatbib]{neurips_2021}

\usepackage[utf8]{inputenc} 
\usepackage[T1]{fontenc}    
\usepackage{hyperref}       
\usepackage{url}            
\usepackage{booktabs}       
\usepackage{amsfonts}       
\usepackage{nicefrac}       
\usepackage{microtype}      
\usepackage{xcolor}         
\usepackage{graphicx}
\usepackage{amsmath}

\usepackage{multirow}
\usepackage{wrapfig}

\title{A Graph Policy Network Approach for Volt-Var Control in Power Distribution Systems}

\author{%
  Xian Yeow Lee \\
  Iowa State University\\
  Siemens Technology\\
  \texttt{xylee@iastate.edu} \\
   \And
  Soumik Sarkar \\
  Iowa State University\\
  \texttt{soumiks@iastate.edu}
   \AND
  Yubo Wang \\
  Siemens Technology\\
  \texttt{yubo.wang@siemens.com}
}

\begin{document}

\maketitle

\begin{abstract}
Volt-var control (VVC) is the problem of operating power distribution systems within healthy regimes by controlling actuators in power systems. Existing works have mostly adopted the conventional routine of representing the power systems (a graph with tree topology) as vectors to train deep reinforcement learning (RL) policies. We propose a framework that combines RL with graph neural networks and study the benefits and limitations of graph-based policy in the VVC setting. Our results show that graph-based policies converge to the same rewards asymptotically however at a slower rate when compared to vector representation counterpart. We conduct further analysis on the impact of both observations and actions: on the observation end, we examine the robustness of graph-based policy on two typical data acquisition errors in power systems, namely sensor communication failure and measurement misalignment. On the action end, we show that actuators have various impacts on the system, thus using a graph representation induced by power systems topology may not be the optimal choice. In the end, we conduct a case study to demonstrate that the choice of readout function architecture and graph augmentation can further improve training performance and robustness.
\end{abstract}

\section{Introduction} 
Volt-Var Control (VVC) refers to the objective of maintaining the healthy operation of power distribution systems via controlling the voltage (Volt) and reactive power flow (Var). As circuits in power distribution systems usually follow a tree topology, nodes far away from the substation (root node) are usually subject to low voltage magnitude. VVC optimizes the operation of grid assets such as voltage regulators, capacitors, and batteries to minimize power delivery losses as well as to maintain the voltage profile at each bus (node). As such, VVC is one of the critical elements in distribution automation~\cite{borozan2001integrated} and has been studied for decades~\cite{baran1999volt}. VVC has never been a trivial problem to solve. The operational change of any single grid asset may result in change over the whole power distribution systems, as the nodal voltages/currents are connected by the networked Ohm’s law. Researchers have addressed the VVC problem by three different approaches. The first is by sensing the voltage/power profiles near the grid assets and operate the assets based on local observations~\cite{kersting2006distribution}. This is a simple solution to the VVC problem, which may not necessarily need communication links. Though it is still common practice in the field, the performance of such control could sometimes be poor and requires tremendous effort from the engineers to tune the controllers of the grid assets before/during deployments. 

With the development of control and communication technology, researchers have formulated the VVC as an optimal power flow (OPF) problem and leveraged optimization solvers to solve it. However, due to the nature of OPF problems, the resulting optimization problem is non-convex and thus hard to solve. Yang et al.~\cite{yang2016optimal} is an example that uses linearization techniques, and Gan et al.~\cite{gan2014exact} is another example that uses convex relaxations to circumvent these challenges. These techniques work under certain strong assumptions (operating around the operating point for linearization and voltage around nominal values for convex relaxation). Together with the introduction of integer/discrete control variables modeling actuators, the OPF problem can hardly be scaled to medium and large systems. With recent breakthroughs in RL, power systems researchers have attempted to use RL for power systems operations. One of such examples is learning to operate a transmission systems operation in L2RPN competition \cite{marot2021learning}. Though transmission systems are fundamentally different from distribution systems in both network topology  (looped vs. radial)  and typical problem types  (dynamic systems vs. quasi-static systems), RL has shown promising results ~\cite{yoon2020winning} in transmission systems. 

In this work, we follow the third approach of utilizing RL to learn an optimal control strategy for the VVC problem and investigate the feasibility of combining such approaches with recent breakthroughs in graph representations to learn better controllers. Specifically, this study attempts to answer the following question: does providing topology information of the power system via a graph representation and using graph convolutional network instead of dense network policies for VVC beneficial to the 1) sample efficiency/convergence rate of training the control policy, 2) robustness of the trained policy to typical data acquisition errors in power systems, and 3) what is a good representation of the underlying circuit topology given the possible different impact of actuators.

This paper contributes to existing literature by being the first attempt to combine graph-based networks with reinforcement learning for VVC. We demonstrate that this graph-based agent and system representation easily scales to power systems of multiple sizes from small (13Bus) to medium-size systems (8500Nodes), and its performance is competitive with the dense policy counterpart. Additionally, we show that a graph-based policy is much more robust in the presence of missing/noisy observations as compared to a dense-based policy. Furthermore, we show that unlike many other graph representation studies, the actuators in VVC could be using a different topology during message passing compared to the well studied topology induced by physical systems (in our case the circuit). Lastly, we present a case study to investigate the effects of graph augmentation and alternative readout function and show such factors can further improve the performance of the graph-based agent.

\vspace{-7 pt}
\section{Related Works}
\vspace{-7 pt}

We briefly go over and highlight some of the recent literature related to this work. With recent advancements in graph neural networks~\cite{wu2020comprehensive}, there have been multiple works that leverage the expressive power of graph neural networks with reinforcement learning to solve sequential decision-making problems that have inherent graph representations. Some of the applied works include optimizing network routing~\cite{almasan2019deep}, compilers~\cite{zhou2020transferable}, job scheduling~\cite{mao2019learning} and automatic design of transistor circuit that is transferable~\cite{wang2020gcnrl}. Additionally, the same concept has also been extended applications such as better behavior generation in self-driving applications~\cite{hart2020graph}, learning policies that have zero-shot transfer capabilities in robotics~\cite{wang2018nervenet} and controlling the dynamics of a graph which can be used to represent various real-life scenarios~\cite{pmlr-v139-meirom21a}.

With reference to power systems domains, RL approaches are a well-studied topic, with RL being applied to a multitude of problems from various aspects~\cite{8859593, GLAVIC20176918}.
In the context of VVC, existing studies were mainly focused on various aspects of scaling RL to the challenges specific to the VVC, such as minimizing constraint violations~\cite{8944292} and scaling to combinatorially large actions spaces~\cite{9143169}. Alternatively, researchers have also tackled the VVC problem by formulating it as multi-agent reinforcement learning (MARL) problem and proposing a novel efficient and resilient MARL algorithm~\cite{consensusmarlvvc}. Additionally, a more recent and closely related work in terms of methodology by~\cite{8500nodework} also proposed to combine RL with graph neural networks for power system restoration via a multi-agent formulation. Inspired by these related works, we apply the idea of graph-based RL to solve the volt-var control problem using a graph representation as a complementary extension to the works mentioned above. 

\section{Preliminaries}
\subsection{Volt-var control}
VVC can be formulated as an optimal power flow problem, i.e., optimize for an objective function subject to the physical networked constraints. We represent a power distribution system as a tree graph ($\mathcal{N}, \xi$), where $\mathcal{N}$ is the set of nodes or \textit{buses} and $\xi$ is the set of edges or lines and transformers. 
Let us denote node $i$ as $j$'s parent (radial graph is a tree). In the VVC problem, we consider three different types of actuators, i.e., voltage regulators (reg), capacitors, and batteries. The VVC problem can be described with the following optimization problem~\cite{Farivar2013voltage}. 

\begin{equation}
\begin{split}
\min_{x:\{r,p_{\text{bat}},q_{\text{cap}}\}} &  f_{\text{volt}}(x)+f_{\text{ctrl}}(x)+f_{\text{power}}(x)\\
\text{s.t. }
p_j&=p_{ij}-R_{ij}\ell_{ij}-\sum_{(j,k)\in \xi}p_{jk}+\sum_{m \in j}p_{\text{bat}}^m\\
    q_j&=q_{ij}-X_{ij}\ell_{ij}-\sum_{(j,k)\in \xi}q_{jk}+\sum_{n \in j}q_{\text{cap}}^n\\
    v_j^2&=
    \begin{cases}
    rv_i^2, &\text{if $(i,j)$ is reg} \\
    v_i^2-2(R_{ij}p_{ij}+X_{ij}q_{ij})+(R^2_{ij}+X^2_{ij})\ell_{ij}, & \text{otherwise}
    \end{cases}
    \\
    \ell_{ij}&=(p^2_{ij}+q^2_{ij})/ v_i^2,\\
    r, & \text{ } p_{\text{bat}}, \text{ } q_{\text{cap}} \in \boldsymbol{S}
\end{split}
\label{eq:voltvar}
\end{equation}
The VVC objective is a combination of three losses: $f_{\text{volt}}$ for voltage violation at buses, $f_{\text{ctrl}}$ for control error which prevents actuator wear out by penalizing the actuator status from changing too frequently, and $f_{\text{power}}$ for power loss. $p,q$ are active and reactive power consumed at buses (nodes) or power flow over lines (edges) , $v,\ell$ denote bus voltage magnitude and squared current magnitude, $R,X$ are resistance and reactance. Using $\mathbb{R}$ for real numbers and $\mathbb{Z}$ for integers, $p_{\text{bat}}\in \mathbb{R}$ is the battery power output, $q_{\text{cap}}\in\mathbb{Z}$ denotes the capacitor reactive power output. $r\in \mathbb{Z}$ represents the tapping of voltage regulators. All $\{r, p_{\text{bat}}, q_{\text{cap}}\}$ needs to be operating under their operational constraints captured by set $\boldsymbol{S}$. Capital letters stand for given parameters while lower case letters denotes auxiliary variables, excluding the decision variables. Note that the VVC problem is a time dependent problem, for brevity, we have ignored the time $t$ in all variables. The constraints in Eq.~\eqref{eq:voltvar} have quadratic equality constraints, making any optimization upon it non-convex. 

\subsection{Reinforcement learning}
Based on the previous section, the VVC problem can be formulated as a sequential Markov Decision Problem (MDP) as the power system has no memory and the system's transition into the next state is solely dependent on the control decision and current state. Hence, we can cast the problem as an MDP and solve it via conventional RL approaches, where the objective function is formulated as the reward function in the RL framework. Note that the constraints in the VVC can also be absorbed as part of the reward function or handled explicitly via constrained-RL algorithms~\cite{achiam2017constrained,zhang2020first}.

Formally, let $H$ denote the horizon of an episode and let $f(s_t, a_t)$ denote the underlying environmental dynamics which transitions the system into the next state, $s_{t+1}$, according to the current state, $s_t$ and action $a_t$ based on the physics equation described in the previous section. We parameterized the RL policy, $\pi_{\theta}(s_t)$ using a (graph) neural network, where $a_t \sim \pi_{\theta}(s_t)$. The objective of the RL algorithm is then to optimize parameters of the neural network which maximizes the cumulative reward function, subject to the environment dynamics:
\begin{equation}
\begin{aligned}
& {\max_{\theta}}  \sum_{t=0}^{H} R(s_t,a_{t}, s_{t+1}) & \text{s.t.} & \  s_{t+1} = f(s_{t}, a_{t}), \ a_{t}\sim \pi_{\theta}(s_t)\\
\end{aligned}
\end{equation}
Most RL algorithms typically fall between the spectrum of value-based methods and policy optimization methods. Value-based methods such as Q-learning~\cite{watkins1992q} and its variants aim to learn a function that estimates the value of any state-action pairs and the policy proceeds to take action, which maximizes the estimated cumulative reward. On the other hand, vanilla policy optimization methods, i.e., REINFORCE\cite{sutton2000policy} aims to directly learn a reward-maximizing policy by estimating the gradient of the policy and updating the policy. Finally, actor-critic methods such as A2C~\cite{mnih2016asynchronous} and SAC~\cite{haarnoja2018soft} borrows heavily from both sides to achieve a much better training performance. In this paper, we focus on the PPO algorithm~\cite{schulman2017proximal} with the clipped objective and actor-critic architecture as the baseline for comparing different dense versus graph-based policies due to PPO's ability to scale to multi-discrete action spaces. 

\subsection{Graph neural networks}

Graph neural networks (GNN) are a special class of neural networks designed to learn from graph-structured data by capturing the relationship between different nodes. Recent literature have illustrated that different GNNs architectures can be designed and applied to various inductive and transductive tasks such as node classification, edge prediction, and graph classification. The core idea of GNNs is to learn embeddings using the message-passing mechanism, where the features of a node in the graph are aggregated based on the features of neighboring nodes, with some learnable parameters transforming the messages. Depending on the downstream application at hand (node classification, graph classification, etc.), the learned nodal embeddings are typically further aggregated and/or sent through a readout function (usually a dense layer) that outputs a final prediction. In this work, we focus specifically on Graph Convolutional Networks (GCN)~\cite{kipf2016semi} as a possible representation of $\pi_{\theta}$. Formally, let $h^{i+1}_{v}$ denote the features/embeddings of node $v$ at the next layer and let $\sigma$ denote any arbitrary non-linearity function, $b_i$ a learnable bias and $b_i$, $\phi^{i}$ the learnable weight term of the current layer. Further, let $N(v)$ denote the neighborhood of node $v$, $E_{u,v}$ denote the edge weight between nodes ($u,v$), $C_{u,v}$ some normalization constant and $h^i_{u}$ the features/embeddings of the current layer. Then the nodal embeddings of the $i+1$ layer can be defined as:
\begin{equation}
\begin{aligned}
h^{i+1}_{v} = \sigma(b^{i} + \sum_{u \in N(v)}  \frac{E_{u,v}}{C_{u,v}} \phi^{i} h_{u}^{i})
\end{aligned}
\label{eqn:gcn}
\end{equation}
Note that the number of layers $i$ is also representative of the depth of the node neighborhood we consider, i.e., the number of hops information is propagated along. In the context of using a GNN as an RL policy, we use the final embeddings $h^{i+1}_{v}$ as inputs to a readout function, $\mathcal{R(.)}$ (i.e., mean-pooling of nodal embeddings) to generate the logits to a dense layer which outputs the actions. We further highlight that using other variants of GNNs such as using alternate forms of readout functions, aggregation functions~\cite{hamilton2017inductive} and including attention on edge weights~\cite{velivckovic2018graph} is also possible and is complementary to our approach. 

\vspace{-5 pt}
\section{Methodology}
\vspace{-5 pt}

To answer the question of whether using a graph representation and GNN-based policy for VVC result in better performances, we trained dense- and graph-based PPO agents on the PowerGym environment~\cite{fan2021powergym}, which consists of benchmark systems of various sizes. Specifically, we trained separate policies on four power systems: 13Bus, 34Bus, 123Bus, and 8500Node~\footnote{From the context of power systems, nodes are counted per phase, and a bus may contain up to three phases. Hence, a 13Bus system may have $\geq$ 13 nodes, and the 8500Node system will have $\leq$ 8500 buses}. In these systems, the agent controls different combinations of voltage regulators, capacitors, and batteries to modulate the node voltage across the entire system within a normalized threshold of 0.95 and 1.05V while simultaneously reducing overall control cost and system power loss. 

\textbf{Environment: }The reward function of the environments takes the form of $R_{t} = \alpha_{v}R_{v} + \alpha_{c}R_{c} + \alpha_{p}R_{p}$, where $R_{v}$ denotes the penalty for voltage violations, $R_{c}$ penalty for changing controls, $R_{p}$ penalty for power loss and $\alpha_{v}, \alpha_{c}, \alpha_{p}$ denotes the respective weightage for each penalty term. These terms map back to the three terms in the objective function of Eq.~\eqref{eq:voltvar}. Note that the terms $R_{v}$ and $R_{c}$ may also be conflicting in practice. Minimizing the voltage violation, $R_{v}$ would require frequent operation of voltage regulators, capacitors, and batteries, which would subsequently increase the control cost, $R_{c}$ and vice versa. Thus, it results in a multi-objective RL scenario which would require careful tuning of $\alpha_{v}$ and $\alpha_{c}$. Nevertheless, for simplicity, we used the prescribed reward function and weights presented in the benchmark environment. The observations of the PowerGym environment consist of the minimum phase voltage at every bus as well as the status of the controllable actuators, i.e., voltage regulators, capacitors, and batteries. The actions consist of changing the status of the controls, such as changing the tapping of a voltage regulator, switching the capacitors state between open/close, and charging/discharging the batteries. Since the capacitors have only two switchable states while the batteries and voltage regulators have multiple switchable states, the action space is represented by a multi-discrete action space, where each actuator can be controlled with independent and potentially different numbers of discrete actions. For more details on the environment setup, we defer the reader to the article on PowerGym~\cite{fan2021powergym}.

\textbf{Representation and training: }To compare the training performance of dense versus graph-based policy, we trained the PPO agents on 20,000 episodes, with each episode having a horizon of 24 steps and subjected to different energy load profiles for every episode to introduce variations. For the agent parameterized with the dense policy (Dense-PPO), we concatenated the observation of the environment as a vector input to the policy, as is commonly done in literature. For the agent parameterized by a GCN policy, denoted as Graph-PPO, we constructed a graph representation according to the topology of the physical power systems and assigned every node the features consisting of the node voltage and capacitor, voltage regulator, and battery status. Nodes that do not have capacitors, voltage regulators or batteries are padded with zeros. 
The entire graph representation of the power system's state is then sent as input to the GCN for Graph-PPO. To facilitate fair comparisons, we used the same number of layers and hidden dimensions for both the dense and GCN network to compare the performance on the same system. All experiments were performed on a server with one AMD Ryzen Threadripper 3970X CPU and one Nvidia RTX 3090 GPU. Further details on training hyper-parameters are provided in the supplementary materials. 

\vspace{-5 pt}
\section{Results and Discussions}
\vspace{-5 pt}
\subsection{Training convergence}
\label{sec:training_convergence}

\begin{figure*}[h!]
  \centering
  \includegraphics[width=0.9\linewidth, clip, trim={0in 0in 0in 0in}]{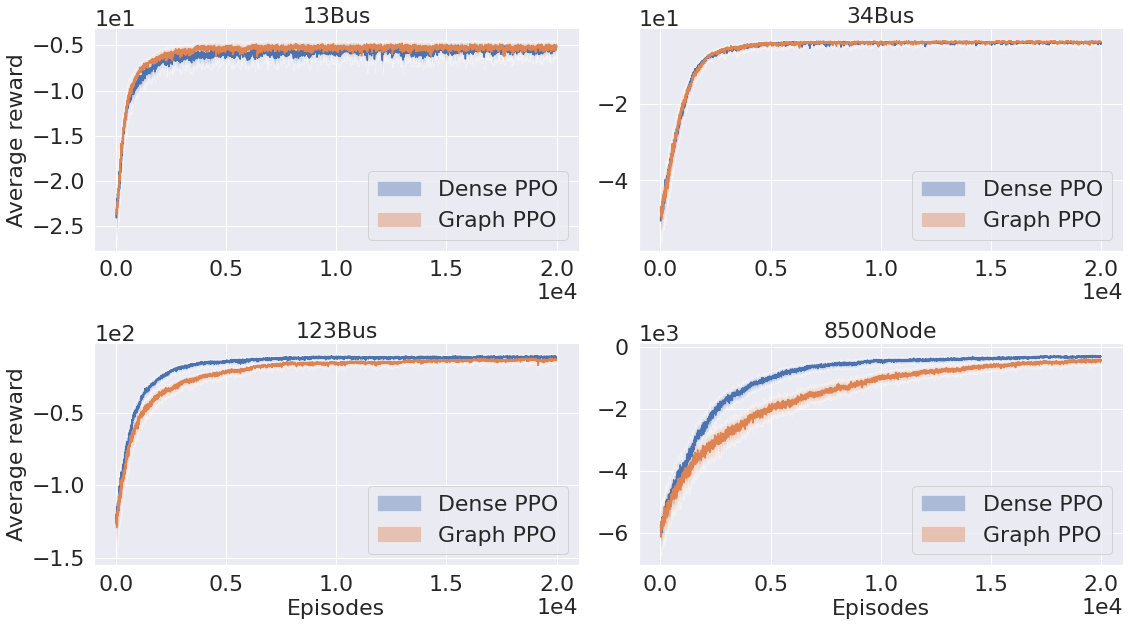}
  \caption{Rewards achieved by Dense-PPO and Graph-PPO over 20,000 training episodes. Rewards shown are averaged over five random seeds. Graph-PPO performs marginally better than Dense-PPO for smaller systems but converges much slower than Dense-PPO for larger systems.}
  \label{fig:train_rewards}
\end{figure*}

We first compare the results of training the RL agent using a dense policy versus a GCN-based policy. Figure~\ref{fig:train_rewards} illustrates the rewards obtain by the RL policy during training for the 13Bus, 34Bus, 123Bus, and 8500Node systems, respectively. There are two keys observations that are worth noting. First, we observe that both Dense-PPO and Graph-PPO are able to converge to similar rewards asymptotically for all four systems. This validates the fact that both policies can learn from their respective representations. The second observation is that for a smaller system (13Bus), Graph-PPO is able to converge to a marginally higher reward than Dense-PPO. Nevertheless, as we increase the size of the system, also corresponding to a larger state representation, we can see that Dense-PPO outperforms Graph-PPO.
\\
\\
From these observations, we conclude that using a graph representation and graph-based policy does not provide any significant benefit in terms of training convergence. In fact, using a graph representation does result in significantly slower training for larger systems (8500Node), which are much closer to the size of power distribution systems in real life. However, this conclusion is subjected to some additional caveats. We hypothesize the reason that Graph-PPO converges slower for larger systems is due to the message passing nature of the graph convolutional layers. Since the number of graph convolutional layers represents the number of hops away message from a node is allowed to propagate to its neighboring nodes, this severely restricts the representation of the system if there is a presence of a node with global effects while the number of graph convolution layers is limited in the GCN. This hypothesis is further validated in subsection~\ref{sec:sensitivity}. In contrast, dense layers in Dense-PPO are essentially able to learn correlations between the embeddings of any nodes, regardless of the connectivity of the physical systems, thus allowing Dense-PPO to learn faster. It is worth highlighting that for the experiments shown, we have followed the convention of most state-of-the-art GCNs, which uses only three to four layers deep network, partially due to the small-world properties~\cite{watts1998collective} in such applications. As such, we have used a three-layer GCN for 13Bus, 34Bus, and 123Bus system and a four-layer GCN for the 8500Node system. On the contrary, a traversal of the 123Bus and 8500Node graph reveals a tree depth of 26 and 272, respectively. Furthermore, it has also been shown that using very deep GCNs can lead to the over-smoothing phenomenon~\cite{chen2020measuring}, where the useful signals in a node's features/embeddings get over-mixed by neighboring nodes.

\vspace{-5 pt}
\subsection{Robustness to typical data acquisition errors}
\vspace{-5 pt}

In power systems, it is not uncommon to have communication failures which results in missing voltage readings (missing observations). Further, voltage reading can be very noisy (noisy observations) due to misalignment of sampling time, i.e., voltage meters measuring different snapshots of the same power system.  We examine the benefits and drawbacks of Graph-PPO versus Dense-PPO from a perspective of the robustness of the learned control policy towards the absence of features or presence of noise in the features. While the general robustness of RL policies to adversarial attacks has been widely studied in literature~\cite{ilahi2020challenges, chen2019adversarial, lee2021query}, we investigate the control policy's robustness towards missing and noisy observations/measurements as they are a realistic setting in the context of power distribution systems. To simulate such settings and test the robustness of the Graph-PPO and Dense-PPO, we designed two experiments. In the first experiment, to simulate communication failure (missing observations), we randomly selected a subset of nodes in the power distribution system and set the voltage values to zero during testing. This resulted in certain values in the input vector being zero for Dense-PPO and certain nodal features being zero in the input graph for Graph-PPO. To simulate measurement misalignment (noisy measurements) in the second experiment, we generated uniform random noise, and added them to the voltage values of randomly sampled subsets of nodes. 
\begin{table}[h!]
    \centering
    \begin{tabular}{l | c c c | c c c}
        \hline
        \hline
        &\multicolumn{3}{c}{Dense-PPO}&\multicolumn{3}{c}{Graph-PPO}\\
        \hline
        System   & 25\% & 50\%  & 75\%  & 25\%  & 50\%  & 75\%  \\
        \hline
        13 Bus   &$-25\pm 3$ &$-75\pm 13$& $-154\pm30$&  $-7\pm3$&$-24\pm4$ &$-45\pm2$ \\
        
        34 Bus   &$-17\pm7$ &$ -54\pm 14$ & $-167\pm25$ & $-5\pm2$&$-13\pm4$ &$-47\pm43$ \\
        
        123 Bus  &$-38\pm16$&$-111\pm28$& $-272\pm31$&  $-2\pm1$&$-10\pm2$ &$-12\pm4$ \\
        
        8500 Bus &$-14\pm4$&$-37\pm9$& $-86\pm3$&  $-5\pm3$&$-5\pm3$ &$-5\pm3$ \\
        \hline
        \hline
    \end{tabular}
    \caption{Values in the table denote the percent difference of rewards from the nominal setting with full observations. Each column represents a scenario with a certain percent of missing node voltage observations due to sensor communication failures.}
    \label{tab:masking}
\end{table}
\vspace{-5 pt}

\textbf{Sensor communication failures :} 
Table~\ref{tab:masking} shows the effects of missing observations for both Dense-PPO and Graph-PPO on all four systems with a different number of node voltage values masked. Each value in the table represents the average percentage decrease in rewards of the agent in the missing observation setting from the nominal setting with full observations available. Each average value was obtained by averaging the drop in performance across five random subsets of nodes and agents trained with different seeds. As observed, though Dense-PPO and Graph-PPO are both negatively affected by missing observations, the effect of missing observations is significantly more severe on Dense-PPO as compared to Graph-PPO. Additionally, we see that the significance of this effect increases as we increase the number of missing observations. In the best-case scenario, Dense-PPO suffers at least 3X worse than Graph-PPO, and in the worst-case scenario, Dense-PPO is 23X worse than Graph-PPO (123Bus with 75\% missing observations). Overall, this demonstrates that Graph-PPO is capable of maintaining a much more robust performance than Dense-PPO in the event of missing observations. Additional visualizations illustrating the performance Dense-PPO and Graph-PPO in the nominal and missing observation settings are presented in the supplementary materials.


\begin{table}[h!]
    \centering
    \begin{tabular}{l | c c c | c c c}
        \hline
        \hline
        &\multicolumn{3}{c}{Dense-PPO}&\multicolumn{3}{c}{Graph-PPO}\\
        \hline
        System   & 25\% & 50\%  & 75\%  & 25\%  & 50\%  & 75\%  \\
        \hline
        13 Bus   &$-8\pm 1$ &$-23\pm 7$& $-43\pm8$&  $-3\pm3$&$-9\pm4$ &$-16\pm5$ \\
        
        34 Bus   &$-6\pm4$ &$ -21\pm 5$ & $-39\pm6$ & $-12\pm3$&$-21\pm5$ &$-28\pm3$ \\
        
        123 Bus  &$-13\pm5$&$-30\pm6$& $-63\pm7$&  $-8\pm4$&$-11\pm6$ &$-13\pm2$ \\
        
        8500 Bus &$-7\pm5$&$-11\pm3$& $-23\pm5$&  $-5\pm3$&$-4\pm3$ &$-4\pm2$ \\
        \hline
        \hline
    \end{tabular}
    \caption{Values in the table denote the percent difference of rewards from the nominal setting with clean observations. Each column represents a scenario with noise added to the voltage values of certain subset of nodes mimicking measurement misalignment. }
    \label{tab:noisy}
\end{table}

\textbf{Measurement misalignment:} Table~\ref{tab:noisy} presents the performance of Dense-PPO and Graph-PPO under misaligned measurements. Comparing the scenario of noisy observations to missing observations, we observe that both Dense-PPO and Graph-PPO are less sensitive to noisy observations than missing observations. Nonetheless, we still observe a similar trend where Graph-PPO maintains a much more robust performance in the presence of noisy observations, especially when a larger percentage of the observations are noisy (i.e., 50\% and 75\% columns in Table~\ref{tab:noisy}).

From the results above, we can affirmatively confirm that Graph-PPO enables a much more robust control policy than Dense-PPO in a situation when voltage readings may be missing or noisy. This occurrence can also be attributed to the message-passing mechanism of the GCN in a Graph-PPO. When voltage observations of any nodes are missing or noisy, the GCN architecture enables the information of neighboring nodes to naturally fill in the missing values or smoothen out the noisy values to generate a much more accurate overall state representation. Another possible cause is that the dense layers of Dense-PPO, in the absence of node-connectivity constraints in the vector representations, have learned spurious correlations among non-connected nodes. This has allowed the Dense-PPO to converge much faster during training but also subsequently caused the policy to overfit to the relative positions of voltage observations in the vector representation to a certain degree. In the following sections, we attempt to further investigate this claim via subsection~\ref{sec:sensitivity} and subsection~\ref{sec:casestudy}.

\vspace{-5 pt}
\subsection{Actuator sensitivity analysis}
\label{sec:sensitivity}
\vspace{-5 pt}

In the results presented above, we have made an assumption that the graphical representation of the power system follows the topology of the physical power system, i.e., only buses that are connected by lines will have an edge connecting the nodes in the graph. This representation can limit the information propagation between nodes when using a limited number of GCN layers. Thus, changing features of one node will have a larger effect on nodes that are directly connected as compared to nodes that are connected yet far away. However, from the power systems perspective, not all the actuators behave the same way. Certain actuators, such as the voltage regulator, have a global effect, while batteries and capacitors have a more local effect. The following section serves to validate the presence both global and local nodes specific to the domain of power transmission system.  

\begin{wrapfigure}{r}{0.42\textwidth}
  \begin{center}
    \includegraphics[width=0.36\textwidth]{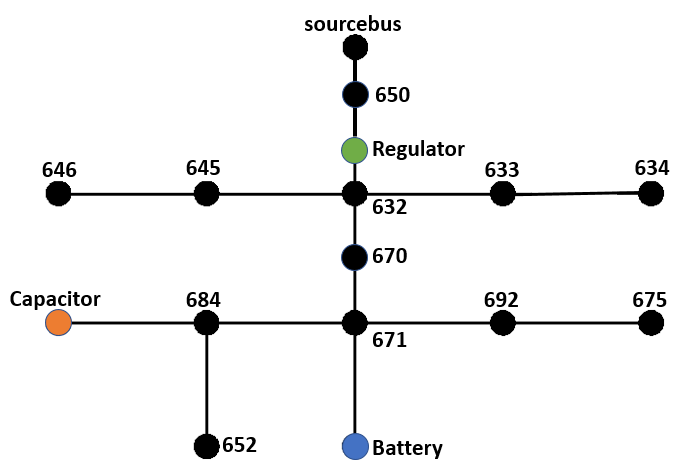}
  \end{center}
  \caption{Topology of 13Bus power distribution system.}
  \label{fig:layout}
\end{wrapfigure}

\begin{figure*}[h!]
  \centering
  \includegraphics[width=1\linewidth, clip, trim={0in 0in 0in 0in}]{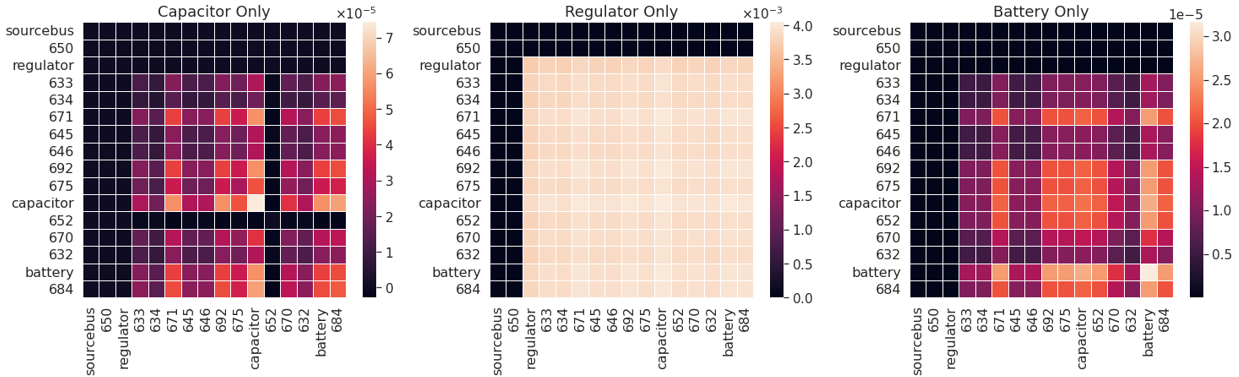}
  \caption{Visualization of co-variance between nodes in 13Bus system when only capacitor, voltage regulator or battery is active. Observe that the effects of capacitors and batteries are local while the effect of the voltage regulator is global, although the voltage regulator may not be directly connected to every node.}
  \label{fig:sensitivity}
\end{figure*}

\textbf{Sensitivity analysis}: We conducted a sensitivity analysis on the power system by running the random policy for a single episode using only one active actuator (while disabling the other actuators) and observed the co-variance of the voltage between the actuator bus and the voltage of all other buses. For easy visualization purposes, we demonstrated this on a 13Bus system, shown in Figure~\ref{fig:layout}. Figure~\ref{fig:sensitivity} illustrates the co-variances between the buses when only the capacitor, voltage regulator, or battery is active. As observed in Figure~\ref{fig:sensitivity}, we can tell that the voltage regulator clearly has a global effect on every bus, even though not all the buses are directly connected to the voltage regulator bus. In contrast, the capacitor and batteries have a more local effect, with the capacitor having a larger neighborhood influence than the battery. 

The property of global versus local effect of physical actuators is a topic that is more oriented towards the domain of engineering controls, and is not present in most of the graph representational learning tasks, which typically assumes the topology used for message passing is induced by the modeling of physical systems. Bearing this observation in mind, we have improved the graph representations in subsection~\ref{sec:casestudy}.

\subsection{123Bus Case Study}
\label{sec:casestudy}
Based on the findings of the previous section, we have validated that global nodes and local nodes do exist in the power systems. However, from the discussions in subsection~\ref{sec:training_convergence}, we also know that the effects of the global nodes aren't properly propagated throughout the entire graph as we are only using three or four graph convolution layers. In this section, we investigate if we can incorporate such knowledge by augmenting the representation of the graph to improve the performance of Graph-PPO. In light of the global and local effects, we also study how different readout functions, $\mathcal{R(.)}$ used in the GCN networks can affect the final performance.

\textbf{Learning with augmented graph representation}: To properly represent the existence of a node with global effects such as the voltage regulator, we augmented the graph representation of the power system in the following way: Given the original graph topology based on the physical system, we traversed the graph from the root (source bus) to the leaves and recorded the child nodes of every voltage regulator encountered during the traversal. We then augmented the graph by adding an edge between the nodes with voltage regulator and all its corresponding child nodes. This effectively mimics the global effects of the voltage regulator node by allowing information to propagate in the GCN layers even if a child node is not directly connected to the voltage regulator node. 

\textbf{Learning with a different readout function}: We also perform a study on the effects of using a different readout functions. As an initial baseline, we have used a mean-pooling readout function to average the embeddings of all the nodes from the final GCN layer in the policy/value networks before passing them to a dense layer to output the control actions (mean-pool readout). This network design has been used in all the results presented in the previous sections. However, averaging the embeddings of all nodes for decision may not be the best approach, especially when the capacitors and batteries only need to observe the neighborhood voltages to exert local control. We tested an alternative readout function where instead of averaging the embeddings of all nodes, we only take the embeddings of the control actuators and stacked them together to form the logits to the final dense layer (local readout). This architecture corresponds to control actions computed based on the embeddings of the capacitor, voltage regulator, and battery nodes. 

\begin{figure*}[h!]
  \centering
  \includegraphics[width=0.6\linewidth, clip, trim={0in 0in 0in 0in}]{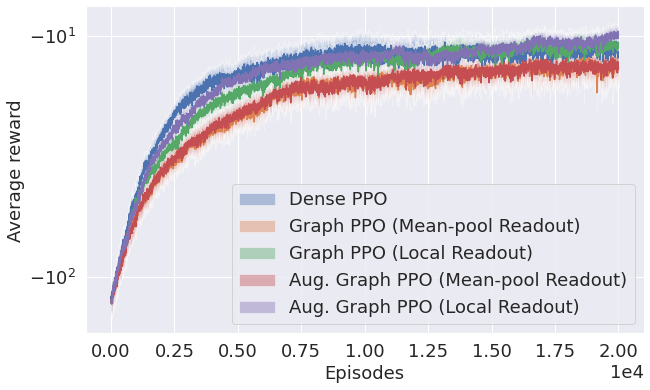}
  \caption{Effects of adding using graph augmentation and localized readout function based only on the nodes with actuators. Observe that augmentation had no effect when paired with mean-pooling readout but using local readout and augmentation gives the best improvement}
  \label{fig:compare_arch}
\end{figure*}

\textbf{Results of graph augmentation and localized readout function}: Figure~\ref{fig:compare_arch} and Table~\ref{tab:casestudy} summarizes the results of exploring the graph augmentation and alternative readout function. From Figure~\ref{fig:compare_arch}, we see that when using a mean-pooling readout, augmenting the graph representation provides no effect as the training performances are similar. In contrast, using a local readout function provides a noticeable increase in performance. Finally, using a local readout function in tandem with graph augmentation further improves the performance almost similar to the Dense-PPO. In terms of robustness during testing (Table~\ref{tab:casestudy}), we observe that the Augmented Graph-PPO with a mean-pooling readout is less robust in terms of communication failure but slightly more robust in terms of measurements misalignment than vanilla Graph-PPO with mean-pooling. However, Augmented Graph-PPO with local readout is consistently the most robust in both settings, while vanilla Graph-PPO using local readout is slightly less robust than Augmented Graph-PPO with local readout. Nevertheless, even the worse performing vanilla Graph-PPO is significantly more robust when compared to using a Dense-PPO, as tabulated in Table~\ref{tab:masking} and~\ref{tab:noisy}. In summary, we have shown that the choice of readout functions can significantly affect the performance of the control policy, and graph augmentation can have added advantages when paired with the proper readout function. 

\begin{table}[h!]
    \centering
  \begin{tabular}{l | c c c c}
        \hline
        \hline
        Communication & Mean-pool & Local & Local Readout & Mean-pool Readout\\
        failure & Readout & with Augmentation & with Augmentation\\
        \hline
        25\% & $-5\pm2$ & $-3\pm3$ & $-6\pm3$ & $-2\pm3$   \\
        
        50\% & $-6\pm5$ &$-6\pm4$ & $-12\pm4$ & $-4\pm4$  \\
        
        75\% & $-12\pm4$ & $-7\pm4$ & $-26\pm4$ & $-2\pm4$ \\
        \hline
        Measurement & Mean-pool & Local & Local Readout & Mean-pool Readout\\
        misalignment& Readout & Readout & with Augmentation & with Augmentation\\
        \hline
        25\% & $-3\pm3$ & $-4\pm3$ & $-3\pm3$ & $-4\pm3$  \\
        
        50\% & $-5\pm3$ & $-4\pm3$ & $-4\pm3$ & $-2\pm3$ \\
        
        75\% &$-8\pm3$ & $-4\pm2$ & $-5\pm2$& $-3\pm2$\\
        \hline
        \hline
    \end{tabular}
    \caption{Summary of the robustness of the variations of Graph-PPO towards communication failure (missing measurements) and measurement misalignment (noisy measurements). Augmented Graph-PPO with local readout provides the best robustness.}
    \label{tab:casestudy}
\end{table}

\section{Conclusion and future work}
This work presents the first attempt of using a graph representation to train a graph-based deep reinforcement learning policy for volt-var control in power distribution systems. We investigated the benefits and drawbacks of learning an RL-based strategy using a graph representation and graph-based policy (Graph-PPO) instead of the conventional vector representation with a dense policy (Dense-PPO). Our results reveal that Graph-PPO can achieve the same rewards as Dense-PPO asymptotically. However, Dense-PPO does converge faster than Graph-PPO for larger systems (i.e., larger graphs). The second result of this work is that while Graph-PPO is slower to train, using a graph representation along with the graph-based policy provides a much more robust control in the event of sensor communication failure and measurement misalignment. In addition, our case study demonstrates that augmentation of the graph representation to account for the presence of nodes with global effects and using an alternative readout function based only on local node embeddings can further improve Graph-PPO's training performance and robustness. Future directions of this work include incorporating advanced graph neural networks architectures as part of the policy representation, extending this framework to more complex settings such as larger and heterogeneous graphs (power systems coupled with communication networks), and investigating the transferability of these graph-based controllers to different power systems topologies.

\newpage
\bibliographystyle{unsrt}
\bibliography{references.bib}

\begin{thebibliography}{10}

\bibitem{borozan2001integrated}
Vesna Borozan, Mesut~E Baran, and Damir Novosel.
\newblock Integrated volt/var control in distribution systems.
\newblock In {\em 2001 IEEE Power Engineering Society Winter Meeting.
  Conference Proceedings (Cat. No. 01CH37194)}, volume~3, pages 1485--1490.
  IEEE, 2001.

\bibitem{baran1999volt}
Mesut~E Baran and Ming-Yung Hsu.
\newblock Volt/var control at distribution substations.
\newblock {\em IEEE Transactions on Power Systems}, 14(1):312--318, 1999.

\bibitem{kersting2006distribution}
William~H Kersting.
\newblock {\em Distribution system modeling and analysis}.
\newblock CRC press, 2006.

\bibitem{yang2016optimal}
Zhifang Yang, Haiwang Zhong, Qing Xia, Anjan Bose, and Chongqing Kang.
\newblock Optimal power flow based on successive linear approximation of power
  flow equations.
\newblock {\em IET Generation, Transmission \& Distribution},
  10(14):3654--3662, 2016.

\bibitem{gan2014exact}
Lingwen Gan, Na~Li, Ufuk Topcu, and Steven~H Low.
\newblock Exact convex relaxation of optimal power flow in radial networks.
\newblock {\em IEEE Transactions on Automatic Control}, 60(1):72--87, 2014.

\bibitem{marot2021learning}
Antoine Marot, Benjamin Donnot, Gabriel Dulac-Arnold, Adrian Kelly, A{\"\i}dan
  O'Sullivan, Jan Viebahn, Mariette Awad, Isabelle Guyon, Patrick Panciatici,
  and Camilo Romero.
\newblock Learning to run a power network challenge: a retrospective analysis.
\newblock {\em arXiv preprint arXiv:2103.03104}, 2021.

\bibitem{yoon2020winning}
Deunsol Yoon, Sunghoon Hong, Byung-Jun Lee, and Kee-Eung Kim.
\newblock Winning the l2rpn challenge: Power grid management via semi-markov
  afterstate actor-critic.
\newblock In {\em International Conference on Learning Representations}, 2020.

\bibitem{wu2020comprehensive}
Zonghan Wu, Shirui Pan, Fengwen Chen, Guodong Long, Chengqi Zhang, and S~Yu
  Philip.
\newblock A comprehensive survey on graph neural networks.
\newblock {\em IEEE transactions on neural networks and learning systems},
  32(1):4--24, 2020.

\bibitem{almasan2019deep}
Paul Almasan, Jos{\'e} Su{\'a}rez-Varela, Arnau Badia-Sampera, Krzysztof Rusek,
  Pere Barlet-Ros, and Albert Cabellos-Aparicio.
\newblock Deep reinforcement learning meets graph neural networks: Exploring a
  routing optimization use case.
\newblock {\em arXiv preprint arXiv:1910.07421}, 2019.

\bibitem{zhou2020transferable}
Yanqi Zhou, Sudip Roy, Amirali Abdolrashidi, Daniel Wong, Peter Ma, Qiumin Xu,
  Hanxiao Liu, Phitchaya~Mangpo Phothilimthana, Shen Wang, Anna Goldie, et~al.
\newblock Transferable graph optimizers for ml compilers.
\newblock {\em arXiv preprint arXiv:2010.12438}, 2020.

\bibitem{mao2019learning}
Hongzi Mao, Malte Schwarzkopf, Shaileshh~Bojja Venkatakrishnan, Zili Meng, and
  Mohammad Alizadeh.
\newblock Learning scheduling algorithms for data processing clusters.
\newblock In {\em Proceedings of the ACM Special Interest Group on Data
  Communication}, pages 270--288, 2019.

\bibitem{wang2020gcnrl}
Hanrui Wang, Kuan Wang, Jiacheng Yang, Linxiao Shen, Nan Sun, Hae-Seung Lee,
  and Song Han.
\newblock Gcn-rl circuit designer: Transferable transistor sizing with graph
  neural networks and reinforcement learning.
\newblock In {\em The 57th Design Automation Conference (DAC)}, 2020.

\bibitem{hart2020graph}
Patrick Hart and Alois Knoll.
\newblock Graph neural networks and reinforcement learning for behavior
  generation in semantic environments.
\newblock In {\em 2020 IEEE Intelligent Vehicles Symposium (IV)}, pages
  1589--1594. IEEE, 2020.

\bibitem{wang2018nervenet}
Tingwu Wang, Renjie Liao, Jimmy Ba, and Sanja Fidler.
\newblock Nervenet: Learning structured policy with graph neural networks.
\newblock In {\em International Conference on Learning Representations}, 2018.

\bibitem{pmlr-v139-meirom21a}
Eli Meirom, Haggai Maron, Shie Mannor, and Gal Chechik.
\newblock Controlling graph dynamics with reinforcement learning and graph
  neural networks.
\newblock In Marina Meila and Tong Zhang, editors, {\em Proceedings of the 38th
  International Conference on Machine Learning}, volume 139 of {\em Proceedings
  of Machine Learning Research}, pages 7565--7577. PMLR, 18--24 Jul 2021.

\bibitem{8859593}
Zidong Zhang, Dongxia Zhang, and Robert~C. Qiu.
\newblock Deep reinforcement learning for power system applications: An
  overview.
\newblock {\em CSEE Journal of Power and Energy Systems}, 6(1):213--225, 2020.

\bibitem{GLAVIC20176918}
Mevludin Glavic, Raphaël Fonteneau, and Damien Ernst.
\newblock Reinforcement learning for electric power system decision and
  control: Past considerations and perspectives.
\newblock {\em IFAC-PapersOnLine}, 50(1):6918--6927, 2017.
\newblock 20th IFAC World Congress.

\bibitem{8944292}
Wei Wang, Nanpeng Yu, Yuanqi Gao, and Jie Shi.
\newblock Safe off-policy deep reinforcement learning algorithm for volt-var
  control in power distribution systems.
\newblock {\em IEEE Transactions on Smart Grid}, 11(4):3008--3018, 2020.

\bibitem{9143169}
Ying Zhang, Xinan Wang, Jianhui Wang, and Yingchen Zhang.
\newblock Deep reinforcement learning based volt-var optimization in smart
  distribution systems.
\newblock {\em IEEE Transactions on Smart Grid}, 12(1):361--371, 2021.

\bibitem{consensusmarlvvc}
Yuanqi Gao, Wei Wang, and Nanpeng Yu.
\newblock Consensus multi-agent reinforcement learning for volt-var control in
  power distribution networks.
\newblock {\em IEEE Transactions on Smart Grid}, 12(4):3594--3604, 2021.

\bibitem{8500nodework}
Tianqiao Zhao and Jianhui Wang.
\newblock Learning sequential distribution system restoration via
  graph-reinforcement learning.
\newblock {\em IEEE Transactions on Power Systems}, pages 1--1, 2021.

\bibitem{Farivar2013voltage}
Masoud Farivar, Lijun Chen, and Steven Low.
\newblock Equilibrium and dynamics of local voltage control in distribution
  systems.
\newblock In {\em 52nd IEEE Conference on Decision and Control}, pages
  4329--4334, 2013.

\bibitem{achiam2017constrained}
Joshua Achiam, David Held, Aviv Tamar, and Pieter Abbeel.
\newblock Constrained policy optimization.
\newblock In {\em International Conference on Machine Learning}, pages 22--31.
  PMLR, 2017.

\bibitem{zhang2020first}
Yiming Zhang, Quan Vuong, and Keith Ross.
\newblock First order constrained optimization in policy space.
\newblock {\em Advances in Neural Information Processing Systems}, 33, 2020.

\bibitem{watkins1992q}
Christopher~JCH Watkins and Peter Dayan.
\newblock Q-learning.
\newblock {\em Machine learning}, 8(3-4):279--292, 1992.

\bibitem{sutton2000policy}
Richard~S Sutton, David~A McAllester, Satinder~P Singh, and Yishay Mansour.
\newblock Policy gradient methods for reinforcement learning with function
  approximation.
\newblock In {\em Advances in neural information processing systems}, pages
  1057--1063, 2000.

\bibitem{mnih2016asynchronous}
Volodymyr Mnih, Adria~Puigdomenech Badia, Mehdi Mirza, Alex Graves, Timothy
  Lillicrap, Tim Harley, David Silver, and Koray Kavukcuoglu.
\newblock Asynchronous methods for deep reinforcement learning.
\newblock In {\em International conference on machine learning}, pages
  1928--1937. PMLR, 2016.

\bibitem{haarnoja2018soft}
Tuomas Haarnoja, Aurick Zhou, Pieter Abbeel, and Sergey Levine.
\newblock Soft actor-critic: Off-policy maximum entropy deep reinforcement
  learning with a stochastic actor.
\newblock In {\em International conference on machine learning}, pages
  1861--1870. PMLR, 2018.

\bibitem{schulman2017proximal}
John Schulman, Filip Wolski, Prafulla Dhariwal, Alec Radford, and Oleg Klimov.
\newblock Proximal policy optimization algorithms.
\newblock {\em arXiv preprint arXiv:1707.06347}, 2017.

\bibitem{kipf2016semi}
Thomas~N Kipf and Max Welling.
\newblock Semi-supervised classification with graph convolutional networks.
\newblock {\em arXiv preprint arXiv:1609.02907}, 2016.

\bibitem{hamilton2017inductive}
William~L Hamilton, Rex Ying, and Jure Leskovec.
\newblock Inductive representation learning on large graphs.
\newblock In {\em Proceedings of the 31st International Conference on Neural
  Information Processing Systems}, pages 1025--1035, 2017.

\bibitem{velivckovic2018graph}
Petar Veli{\v{c}}kovi{\'c}, Guillem Cucurull, Arantxa Casanova, Adriana Romero,
  Pietro Li{\`o}, and Yoshua Bengio.
\newblock Graph attention networks.
\newblock In {\em International Conference on Learning Representations}, 2018.

\bibitem{fan2021powergym}
Ting-Han Fan, Xian~Yeow Lee, and Yubo Wang.
\newblock Powergym: A reinforcement learning environment for volt-var control
  in power distribution systems.
\newblock {\em arXiv preprint arXiv:2109.03970}, 2021.

\bibitem{watts1998collective}
Duncan~J Watts and Steven~H Strogatz.
\newblock Collective dynamics of ‘small-world’networks.
\newblock {\em nature}, 393(6684):440--442, 1998.

\bibitem{chen2020measuring}
Deli Chen, Yankai Lin, Wei Li, Peng Li, Jie Zhou, and Xu~Sun.
\newblock Measuring and relieving the over-smoothing problem for graph neural
  networks from the topological view.
\newblock In {\em Proceedings of the AAAI Conference on Artificial
  Intelligence}, volume~34, pages 3438--3445, 2020.

\bibitem{ilahi2020challenges}
Inaam Ilahi, Muhammad Usama, Junaid Qadir, Muhammad~Umar Janjua, Ala Al-Fuqaha,
  Dinh~Thai Hoang, and Dusit Niyato.
\newblock Challenges and countermeasures for adversarial attacks on deep
  reinforcement learning.
\newblock {\em arXiv preprint arXiv:2001.09684}, 2020.

\bibitem{chen2019adversarial}
Tong Chen, Jiqiang Liu, Yingxiao Xiang, Wenjia Niu, Endong Tong, and Zhen Han.
\newblock Adversarial attack and defense in reinforcement learning-from ai
  security view.
\newblock {\em Cybersecurity}, 2(1):1--22, 2019.

\bibitem{lee2021query}
Xian~Yeow Lee, Yasaman Esfandiari, Kai~Liang Tan, and Soumik Sarkar.
\newblock Query-based targeted action-space adversarial policies on deep
  reinforcement learning agents.
\newblock In {\em Proceedings of the ACM/IEEE 12th International Conference on
  Cyber-Physical Systems}, pages 87--97, 2021.

\end{thebibliography}

\end{document}